\documentclass{article} 
\usepackage{iclr2021_conference,times}

\usepackage{amsmath,amsfonts,bm}

\def\eqref#1{equation~\ref{#1}}

\def\1{\bm{1}}

\DeclareMathAlphabet{\mathsfit}{\encodingdefault}{\sfdefault}{m}{sl}
\SetMathAlphabet{\mathsfit}{bold}{\encodingdefault}{\sfdefault}{bx}{n}

\usepackage[hidelinks]{hyperref}
\usepackage{url}
\usepackage{subfig}
\usepackage{graphicx}
\usepackage{listings}
\usepackage{paralist}
\usepackage{amssymb}
\usepackage{amsmath}
\usepackage{tikz}
\usetikzlibrary{calc}

\newcommand{\bfr}{\mathbf{r}}

\newcommand{\bfx}{\mathbf{x}}

\newcommand{\hath}{\hat{h}}
\newcommand{\bfu}{\mathbf{u}}
\newcommand{\bfs}{\mathbf{s}}

\newcommand{\bR}{\mathbb{R}}

\newcommand{\bP}{\mathbf{P}}
\newcommand{\bS}{\mathbf{S}}

\lstdefinestyle{customc}{
  belowcaptionskip=1\baselineskip,
  breaklines=true,
  frame=L,
  xleftmargin=\parindent,
  language=C,
  showstringspaces=false,
  basicstyle=\footnotesize \ttfamily,
  keywordstyle=\bfseries\color{green!40!black},
  commentstyle=\itshape\color{purple!40!black},
  identifierstyle=\color{blue},
  stringstyle=\color{orange},
}
\DeclareCaptionFont{white}{ \color{white} }
\DeclareCaptionFormat{listing}{
  \colorbox[cmyk]{0.43, 0.35, 0.35,0.01 }{
    \parbox{\textwidth}{\hspace{15pt}#1#2#3}
  }
}

\title{Handling Long-Tail Queries with Slice-Aware Conversational Systems}

\author{Cheng Wang, Sun Kim, Taiwoo Park, Sajal Choudhary \\
Amazon Alexa AI\\
\texttt{\{cwngam, kimzs, parktaiw, sajalc\}@amazon.com} \\

\AND Sunghyun Park, Young-Bum Kim, Ruhi Sarikaya,  Sungjin Lee \\
Amazon Alexa AI\\
\texttt{\{sunghyu, youngbum, rsarikay, sungjinl\}@amazon.com} \\
}

\iclrfinalcopy 
\begin{document}

\maketitle

\begin{abstract}
We have been witnessing the usefulness of conversational AI systems such as Siri and Alexa, directly impacting our daily lives. These systems normally rely on machine learning models evolving over time to provide quality user experience. However, the development and improvement of the models are challenging because they need to support both high (head) and low (tail) usage scenarios, requiring fine-grained modeling strategies for specific data subsets or slices. In this paper, we explore the recent concept of slice-based learning (SBL)~\citep{chen2019slice} to improve our baseline conversational skill routing system on the tail yet critical query traffic. We first define a set of labeling functions to generate weak supervision data for the tail intents. We then extend the baseline model towards a slice-aware architecture, which monitors and improves the model performance on the selected tail intents. Applied to de-identified live traffic from a commercial conversational AI system, our experiments show that the slice-aware model is beneficial in improving model performance for the tail intents while maintaining the overall performance.
\end{abstract} 

\section{Introduction}
Conversational AI systems such as Google Assistant, Amazon Alexa, Apple Siri and Microsoft Cortana have become more prevalent in recent years~\citep{sarikaya2017technology}. One of the key techniques in those systems is to employ machine learning (ML) models to route a user's spoken utterance to the most appropriate skill that can fulfill the request. This requires the models to first capture the semantic meaning of the request, which typically involves assigning the utterance query to the candidate domain, intent, and slots~\citep{el2014extending}. For example, ``Play Frozen" can be interpreted with \textit{Music} as the domain, \textit{Play Music} as the intent, and \textit{Album Name:Frozen} as the slot key and value. Then, the models can route the request to a specific skill, which is an application that actually executes to deliver an experience~\citep{li2021neural}. For commercial conversational AI systems, there usually exists a large-scale dataset of user requests with ground-truth semantic interpretations and skills (e.g., through manual annotations and hand-crafted rules or heuristics). Along with various contextual signals, it is possible to train ML models (e.g., deep neural networks) with high predictive accuracy in routing a user request to the most appropriate skill, which then can continue to optimize towards better user experience through implicit or explicit user feedback~\citep{park2020scalable}.

Nevertheless, developing such ML models or improving existing ones towards better user experience is still challenging. One hurdle is the imbalance in the distribution of the user queries with a long tail in terms of traffic volume. This often makes it difficult for the ML models to learn the patterns from the long-tail queries, some of which could be for critical features. 
Several approaches have been proposed to address such imbalance issue~\citep{smith2014instance,he2008adasyn,chawla2002smote}. However, they are mainly based on applying reverse-discriminative sampling strategies, for example, over-sampling minority and/or under-sampling majority. The sampling methods are usually insufficient in inspecting and improving model performance on pre-defined data subgroups.  

In this work, we focus on the problem of imbalanced queries, specifically on tail but critical intents, in the context of the recently proposed slice-based learning (SBL)~\citep{chen2019slice}. SBL is a novel programming model that sits on top of ML systems. The approach first inspects particular data subsets~\citep{ratner2019role}, which are called slices, and it improves the ML model performance on those slices. While the capability of monitoring specific slices is added to a pre-trained ML model (which is termed the \textit{backbone} model), the approach has shown that overall performance across the whole traffic is comparable to those without SBL. Motivated by this idea, we propose to adopt the SBL concept to our baseline skill routing approach (we term the baseline model as $\bP$; please refer to Sec. \ref{sec:backbone} for details) to improve its performance on tail yet critical intent queries while keeping the overall performance intact. First, we define slice functions (i.e., labeling functions) to specify the intents that we want to monitor. A pre-trained $\bP$ is used as a backbone model for extracting the representation for each query. Then, we extend $\bP$ to a slice-aware architecture, which learns to attend to the tail intent slices of interest.

We perform two experiments using a large-scale dataset with de-identified customer queries. First, we examine the attention mechanism in the extended model $\bP$ with SBL. In particular, we test two attention weight functions with different temperature parameters in computing the probability distribution over tail intent slices. Second, we compare SBL to an upsampling method in $\bP$ for handling tail intents. Our experiments demonstrate that SBL is able to effectively improve the ML model performance on tail intent slices as compared to the upsampling approach, while maintaining the overall performance.

We describe the related work in Section \ref{sec:related_work}. In Section \ref{sec:method}, we explain the baseline skill routing model, $\bP$, and then elaborate how to extend it to a slice-aware architecture. The experiment results are reported in Section \ref{sec:exp}, and in Section \ref{sec:dis}, we discuss the advantages and potential limitations of applying SBL in our use case. We conclude this work in Section \ref{sec:con}.

\section{Related Work}
\label{sec:related_work}

\subsection{Slice-Based Learning}

Slice-based learning (SBL)~\citep{chen2019slice} is a novel programming model that is proposed to improve ML models on critical data slices without hurting overall performance. A core idea of SBL is to represent a sample differently depending on the data subset or slice to which it belongs. It defines and leverages slice functions, i.e.,  pre-defined labeling functions, to generate weak supervision data for learning slice-aware representations. For instance, in computer vision (CV) applications, a developer can define object detection functions to detect whether an image contains a bicycle or not. In natural language understanding (NLP) applications, a developer can define intent-specific labeling functions such as for \textit{Play Music} intent. SBL exhibits better performance than a mixture of experts~\citep{jacobs1991adaptive} and multi-task learning~\citep{caruana1997multitask}, with reduced run-time cost and parameters~\citep{chen2019slice}. Recently Gustavo et al.~\citep{penha2020slice} have employed the concept of SBL to understand failures of ranking models and identify difficult instances in order to improve ranking performance. Our work applies the idea to improve skill routing performance on low traffic but critical intents in conversational AI systems.

\subsection{Weakly Supervised Learning}

Weakly supervised learning attempts to learn predictive models with noisy and weak supervision data. Typically, there are three types of weak supervision: incomplete supervision, inexact supervision, and inaccurate supervision~\citep{zhou2018brief}. Various weakly supervised ML models are developed in NLP~\citep{medlock2007weakly,huang2014learning,wang2014cross} and in CV~\citep{prest2011weakly,oquab2015object,peyre2017weakly}. Recently, promising approaches have been proposed to generate weak supervision data by programming training data~\citep{ratner2016data}. 
In a large-scale industry setting, weak supervision data are highly desired given that human annotations are costly and time-consuming. Our work relates to weakly supervised learning in terms of inaccurate supervision. We split queries into different groups (slices) by defining labeling functions (slice functions). Each group is assigned with a group identity label. In practice, the slice functions may not perfectly assign labels to input data as mentioned in SBL~\citep{chen2019slice,cabannnes2020structured}.

\subsection{Conversational Skill Routing Models}
In conversational AI systems, a skill refers to the application that actually executes on a user query or request to deliver an experience, such as playing a song or answering a question. The skills often comprise both first-party and third-party applications~\citep{li2021neural}. The skill routing is a mechanism that maps users' queries, given contextual information such as semantic interpretations and device types, to an appropriate application. The routing decision is usually determined by an ML model that is separate from typical natural language understanding (NLU) models for domain, intent, slot parsing. Please refer to section~\ref{sec:backbone} for more details.

\section{Slice-Aware Conversational Skill Routing Models}
\label{sec:method}

This section explains our skill routing model (backbone model) and then explains how we extend the backbone model to a slice-aware architecture by adapting the concept from SBL~\citep{chen2019slice}.

\begin{figure}[!htb]
\centering
\subfloat{{\includegraphics[width=0.85\textwidth]{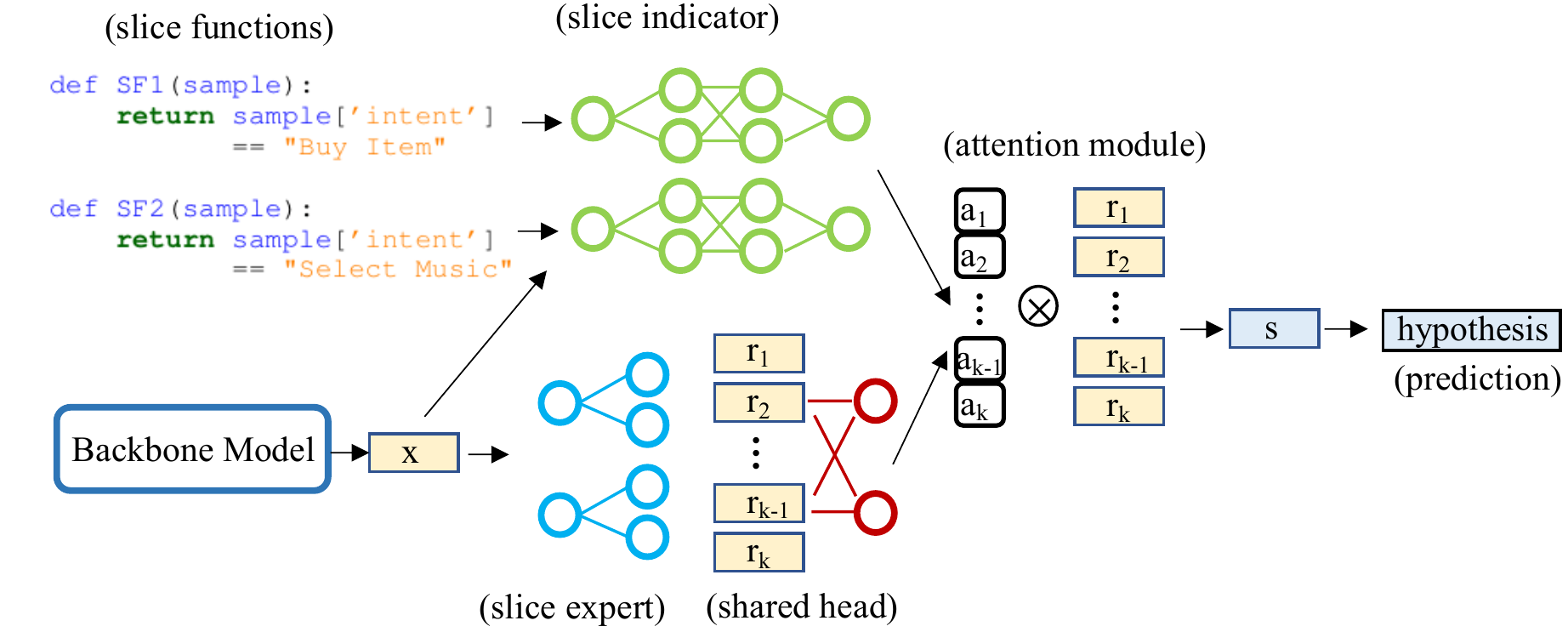}  }}%
\caption{\label{fig:hyprank_slice} The slice-aware conversational skill routing model architecture for handling low traffic but critical intents. It consists of six components: (1) \textit{slice functions} define tail intent slices that we want to monitor; (2) \textit{backbone model} is our pre-trained skill routing model $\bP$ that is used for feature extraction; (3) \textit{slice indicators} are membership functions to predict if a sample query belongs to a tail slice; (4) \textit{slice experts} aim to learn slice-specific representations; (5) \textit{shared head} is the base task predictive layer across experts; (6) An \textit{attention module} is used to re-weight the slice-specific representations $\bfr$ and form a slice-aware representation $\bfs$. Finally, the learned $\bfs$ is used to predict a final hypothesis (associated skill). The predicted hypothesis is used to serve a user query.}
\end{figure}  

\subsection{Backbone Model}
\label{sec:backbone}
We take our baseline skill routing approach ($\bP$) as the backbone model and aim to make it a slice-aware architecture. $\bP$ is a skill routing model, which takes in a list of routing candidates to select the most appropriate one. Each routing candidate is represented as a hypothesis with various contextual signals, such as utterance text, device type, semantic interpretation, and associated skill. While some contextual signals are common across all hypotheses, some are unique due to the presence of multiple competing semantic interpretations and skill-specific context. The core component of $\bP$ consists of attention-based bi-directional LSTMs~\citep{hochreiter1997long,graves2005framewise} with fully connected layers on top of it. Formally, Let $X$ be the set of query signals (e.g., utterance text, semantic interpretations, device type, etc.), $H=\{h_1,...,h_n\}$ be the hypothesis list and $h_g \in H, g=[1, n]$ be the ground-truth hypothesis. The learning objective is to minimize the binary cross entropy: 

\begin{equation}
\zeta_{base}=\mathcal{L}_{bce}(\pi(\mathcal{M}(X, H)), h_g), 
\end{equation}

where $\pi$ is a linear predictive layer which outputs a prediction over hypotheses $\hat{H}=\{\hath_1,...,\hath_n\}$, and $\mathcal{M}$ is a set of multiple neural network layers, which extract the representation $\bfx \in \bR^{n \times d}$ for a given ($X, H$) pair, i.e., $\bfx = \mathcal{M}(X,H)$.

To evaluate the effectiveness of trained $\bP$, we define offline evaluation metric called replication accuracy (RA):

\begin{equation}
\label{eq:ra}
RA(\mathcal{D}_{test}) = \sum\nolimits_{(X, H, h_g)\in\mathcal{D}_{test}} \frac{\mathbb{I}(\hat{h}_g=h_g)}{|\mathcal{D}_{test}|}. 
\end{equation}

The replication accuracy measures how effectively the trained model $\bP$ replicates the current skill routing behavior in production which is a combination of ML model and rules. Though $\bP$ achieves high performance, replicating most of heuristic patterns, it suffers from low RA in low-volume traffic, i.e., the tail user queries. We later introduce how we extend $\bP$ with a slice-aware component.

\subsection{Slice-Aware Architecture}
As presented in Figure~\ref{fig:hyprank_slice}, a slice-aware architecture consists of several components.

\noindent \textbf{Slice Function}. We first define slice (or labeling) functions to slice user queries according to intent (e.g., ``Buy Book"). The selected intents have a small number of query instances, making the model $\bP$ difficult to learn data patterns from tail intents. Each sample is assigned a slice label  $\gamma \in [0, 1]$ in $\{\gamma_1, \gamma_2,...,\gamma_k\}$ for supervision. $s_1$ is the base slice, and $s_2$ to $s_k$ are the tail slices.

\begin{table}
\centering
\begin{tabular}{ll}
\begin{lstlisting}
def intent_based_slice_function_1(sample):
    return sample['intent'] == "Buy Item"

def intent_based_slice_function_2(sample):
    return sample['intent'] == "Select Music"

def intent_based_slice_function_3(sample):
    return sample['intent'] == "Buy Book"
\end{lstlisting}
\end{tabular}
\caption{The slice functions (SFs) which split user queries into multiple data slices according to the pre-defined tail intents. The non-tail intents are in a base slice. Note SFs are only available at training stage for generating weak supervision labels. At inference stage, SFs will not be applied.}
\label{tab:slice}
\vspace{-5mm}
\end{table}

\noindent \textbf{Slice Indicator}. For each tail intent slice, a slice indicator (membership function) is learned to indicate whether a sample belongs to this particular slice or not. For a given representation $\bfx \in \bR^{n \times d}$ from the backbone model, we learn $u_i = f_i(\bfx;\mathbf{w}^f_i)$, $\mathbf{w}^f_i\in \mathcal{R}^{d\times 1}$, $i\in\{1,..,k\}$ that maps $\bfx$ to $\bfu=\{u_1,...,u_k\}$. $f_i$ is trained with $\{\bfx,\gamma\}$ pairs with the binary cross entropy $\zeta_{ind}=\sum_{i}^k \mathcal{L}_{bce}(\bfu_i,\gamma_i)$. 

\noindent \textbf{Slice Expert}. For each tail intent slice, a slice expert $g_i(\bfx;\mathbf{w}^g_i)$, $\mathbf{w}^g_i \in \mathcal{R}^{d\times d}$ is used to learn a mapping from $\bfx \in \bR^{n \times d}$ to a slice vector $ r_i \in \mathcal{R}^{d}$ with the samples only belonging to the tail slice. Followed by a \textbf{shared head}, which is shared across all experts and maps $r_i$ to a prediction $\hath=\varphi(r_i; \mathbf{w}_s)$, $g_i$ and $\varphi$ are learned on the base (original) task with ground-truth label $h_g$ by $\zeta_{exp}=\sum_{i}^k\gamma_i  \mathcal{L}_{bce}(\hath, h_g)$. 

\noindent \textbf{Attention Module}. The attention module decides how to pay special attention to the monitored slices. The distribution over slices (or attention weights) are computed based on stacked $k$ membership likelihood $P \in \bR^k$ and stacked $k$ experts’ prediction confidence $Q \in \bR^{k\times c}$ as described in ~\citep{chen2019slice}:

\begin{equation}
a2=\textsc{softmax}(P+\left | Q \right |).
\end{equation}

Note, the above equation is used when $c=1$ (i.e., binary classification). As our task is a multi-class classification task where $c \geq 2$, we use an additional linear layer to transform $Q \in \bR^{k\times c}$ to $\phi(Q) \in \bR^{k}$.
Finally, we experiment with the following different ways to compute attention weights, i.e., slice distribution:

\begin{align}
\label{eq:m}
a1=\textsc{softmax}(P/\tau)\\
\label{eq:m_e}
a2=\textsc{softmax}([P+\left | \phi(Q) \right |]/\tau).
\end{align}

In Eq. \ref{eq:m}, we only use the output of the indicator function (membership likelihood) in computing attention weights. In Eq. \ref{eq:m_e} we use both the membership likelihood and the transformed experts' prediction scores. The $\tau$ is a temperature parameter. In principle, smaller $\tau$ can lead to a more confident slice distribution~\citep{wang2019state, wang2021uncertainty}, hence we aim to examine if a small $\tau$ helps improve the routing performance.

\section{Experiments}
\label{sec:exp}

We evaluate the skill routing model $\bP$ with slice-based learning (SBL) ~\citep{chen2019slice} (we term it as $\bS$) by performing two groups of experiments. First, we test the attention module with different methods of computing the attention weights over slices. Second, we compare the effectiveness of SBL against upsampling -- a commonly used method for handling tail data.

\subsection{Experiment Setup and Implementation Details}

We obtained live traffic from a commercial conversational AI system in production and processed the data so that individual users are not identifiable. We randomly sampled to create an adequately large data set for each training and test dataset. We further split the training set into training and validation sets with a ratio of 9:1. We used the replication accuracy (Eq. \ref{eq:ra}) to measure the model performance.

The existing production model $\bP$ and its extension with SBL were implemented with Pytorch~\citep{NEURIPS2019_9015}. The hidden unit size for slice component was 128. All models were trained on AWS p3.8xlarge instances with Intel Xeon E5-2686 CPUs, 244 GB memory, and 4 NVIDIA Tesla V100 GPUs.  We used Adam ~\citep{kingma2014adam} with a learning rate of 0.001 as the optimizer. Each model was trained with 10 epochs with the batch size of 256. We split the user queries into 21 data slices in total, one base slice and the rest for 20 tail intent slices. For each extracted query representation $\bfx$ for the tail intents, we add a Gaussian noise $\bfx=\bfx+\delta,~\delta \sim \mathcal{N}(0,0.005)$ to augment the tail queries.

\subsection{Experiments on the Attention Mechanisms}

Table \ref{tab:slice_m} shows the absolute score difference in replication accuracy between the baseline model and its SBL extension,
having the baseline model's all-intent accuracy as a reference. As shown in the table, the slice-based approaches maintain the baseline performance overall, but the RA performance is lifted on the monitored tail slices. The best attention mechanism outperforms the baseline by 0.1\% in tail intents' replication accuracy\footnote{Given the large volume of query traffic per day, 0.1\% is still a significant improvement in our system.}. Tuning the temperature parameter between $\tau=0.1$ or $\tau=1.0$ does not significantly improve model performance on the tail intents.

\begin{table}[!htb]
\centering
\begin{tabular}{ccc}  \\ \hline 
Attention Methods & All Intents (\%) & Tail Intents (\%) \\\hline
$\bP$ (baseline model) & $>$99 & --1.45 \\
SBL, Eq. (\ref{eq:m}), $\tau=1.0$ & +0.01 & --1.35 \\
SBL, Eq. (\ref{eq:m_e}), $\tau=1.0$ & +0.01 & --1.36 \\
SBL, Eq. (\ref{eq:m}), $\tau=0.1$ & +0.01 & --1.34 \\
SBL, Eq. (\ref{eq:m_e}), $\tau=0.1$ & +0.01 & --1.38  \\ \hline 
\end{tabular}
\caption{The performance comparison of the baseline model $\bP$ and its SBL extension with different attention weights in replication accuracy. All data points denote the absolute difference from the baseline model's all intents accuracy value.}
\label{tab:slice_m}
\end{table}

\subsection{Comparison between Slice-Based Learning and Upsampling}

As upsampling is a widely used method to alleviate the tail data problem, we compare the performance between SBL and upsampling methods. Note SBL offers an additional advantage for inspecting particular tail data groups which are also critical. We denote the models as the following:
~~\\
\begin{itemize}
  \item $\bP$ is the baseline model that is trained without applying upsampling.
  \item $\bS$ is an extension of $\bP$ (as a backbone model) to be a slice-aware model, which is trained with same training set as $\bP$.
    \item $\bP_{up}$ is the baseline model that is trained with applying upsampling.
  \item $\bS_{up}$ is an extension of $\bP_{up}$ to be a slice-aware model, which is trained with same training set as $\bP_{up}$.
\end{itemize}

All the trained models are evaluated on the same test set. Among the aforementioned attention method choices, Eq. \ref{eq:m} with $\tau=1.0$ is employed for $\bS$ and $\bS_{up}$. Our primary goal is to see whether $\bS$ can improve $\bP_{up}$.

Table \ref{tab:upsample} shows the performance comparison. When comparing $\bP_{up}$ and $\bS$, we can see $\bS$ achieves slightly better performance for all intents. For the monitored tail intents, $\bS$ achieves a slightly higher score as compared to $\bP_{up}$.

\begin{table}[!htb]
\centering
\begin{tabular}{lcccc} \\ \hline 
Models & All Intents (\%) & Tail Intents (\%) \\\hline
$\bP$ & $>$99 & --1.41 \\
$\bP_{up}$ & 0.00 & --1.47 \\
$\bS$ & +0.01 & --1.30 \\
$\bS_{up}$ & +0.01 & --1.37 \\ \hline
\end{tabular}
\caption{Performance comparison between the baseline model and its slice-aware architecture. $\bP$ is the baseline model without upsampling, $\bP_{up}$ is $\bP$ with upsampling. $\bS$ is the slice learning model with $\bP$ as the backbone model, and $\bS_{up}$ is the slice learning model with $\bP_{up}$ as the backbone model. All data points are absolute score difference from the baseline model's all intent accuracy value.}
\label{tab:upsample}
\end{table}

Table \ref{tab:intent_level_ra} presents the absolute RA difference between the baseline and slice-aware models for the monitored 20 tail intents. Comparing $\bS$ and $\bP_{up}$, $\bS$ improves the model performance on 14 tail intents. Compared to $\bP_{up}$, $\bS$ shows the comparable performance lift while effectively suppressing performance drops, for example, intent IDs 2, 3, 6, 15, and 20. As a result, $\bP_{up}$ shows lower performance on 12 intents out of 20 (--2.41\% on average), while $\bS$ did on only 5 intents (--0.21\% on average). This suggests the capability of slice-based learning in treating target intents through the slice-aware representation.

\begin{table}[!htb]
\centering
\begin{tabular}{ccccccc} \hline
Tail Intent ID & $\bP$ & $\bP_{up}$ & $\bS$ & $\bS_{up}$ & Sample Size \\ \hline
1 & $>$99 & --0.03 & 0.00 & --0.02 & Over 10K \\
2 & $>$96 & --0.4 & +0.04 & --0.21 & Over 10K \\
3 & $>$96 & --0.19 & +0.09 & --0.18 & Over 10K \\
4 & $>$72 & +0.07 & +1.98 & +0.96 & Over 10K \\
5 & $>$99 & +0.01 & --0.01 & 0.00 & Over 10K \\
6 & $>$96 & --0.09 & +0.02 & --0.11 & Over 10K \\
7 & $>$96 & +0.03 & +0.02 & --0.04 & Over 10K \\
8 & $>$99 & +0.15 & +0.01 & +0.19 & Over 10K \\
9 & $>$96 & --0.24 & +0.06 & +0.07 & Over 10K \\
10 & $>$96 & +0.08 & --0.03 & 0.00 & Between 1K - 10K \\
11 & $>$99 & 0.00 & 0.00 & 0.00 & Between 1K - 10K \\
12 & $>$96 & +0.55 & --0.13 & +0.46 & Between 1K - 10K \\
13 & $>$96 & +0.36 & +0.42 & +0.53 & Between 1K - 10K \\
14 & $>$93 & --0.39 & --0.14 & --0.42 & Between 1K - 10K \\
15 & $>$93 & --3.29 & --0.73 & --1.46 & Between 1K - 10K \\
16 & $>$96 & --1.2 & 0.00 & --0.93 & Below 1K \\
17 & $>$96 & --0.71 & 0.00 & --0.71 & Below 1K \\
18 & $>$99 & --0.16 & 0.00 & --0.16 & Below 1K \\
19 & $>$99 & --0.96 & 0.00 & --1.15 & Below 1K \\
20 & $>$96 & --21.21 & 0.00 & --18.18 & Below 1K \\
\hline
\end{tabular}
\caption{Score (in \%) differences in RA between the baseline and slice-aware approaches at the intent level. The baseline model's accuracy scores are rounded down to the nearest multiple of 3 percent, while the other models' are absolute score differences from the baseline ones. We denote each intent with their IDs. Sample Size is the number of random instances used for testing.}
\label{tab:intent_level_ra}
\end{table}

\section{Discussion}
\label{sec:dis}

In our experiments, we have shown the effectiveness of SBL in terms of improving model performance on tail intent slices. It is beneficial to have ML models which are slice-aware, particularly when we want to inspect some specific and critical but low-traffic instances. Although the overall performance gain of slice-aware approach compared to the upsampling was marginal, it is worthwhile to note that the slice-aware approach was able to lift up the replication accuracy for more number of tail intents while minimizing unexpected performance degradation that was more noticeable in the upsampling approach. This result implies that the slice-aware approach has more potential in stably and evenly supporting tail intents.

On the other hand, we also note a potential limitation of SBL in the case of addressing tail intents in the industry setting. As we increase the number of tail intents, for instance to 200 intents, the model's complexity increases as well, given that an indicator function and an expert head are needed for each slice. However, this does not necessarily diminish the value of the slice-aware architecture, as the upsampling method offers no chance for us to inspect and analyze model failures on particular slices. Further studies are necessary to employ and fine-tune the slice-based approach to serve tail traffic in a cost-effective way.

\section{Conclusion}
\label{sec:con}

In this work, we applied and implemented the concept of slice-based learning to our skill routing model for a large-scale commercial conversational AI system. To enable the existing model to pay extra attention to selected tail intents, we tested different ways of computing slice distribution by using membership likelihood and experts' prediction confidence scores. Our experiments show that the slice-based learning can effectively and evenly improve model performance on tail intents while  maintaining overall performance. We also compared the slice learning method against upsampling in terms of handling tail intents. The results suggest that slice-based learning outperforms upsampling by a small margin, while more evenly uplifting tail intents' performance. A potential future work would be to explore how to adapt SBL to monitor a large number of slices with minimum model and runtime complexity.
\clearpage
\newpage

\bibliography{iclr2021_conference}
\bibliographystyle{iclr2021_conference}


\end{document}